\documentclass[11pt,a4paper]{article}
\usepackage[hyperref]{acl2018}
\usepackage{times}
\usepackage{latexsym}
\usepackage{amsmath}
\usepackage{graphicx}
\usepackage{url}
\usepackage{amssymb}
\usepackage{bm}
\usepackage{enumitem}

\aclfinalcopy 
\def\aclpaperid{298} 

\DeclareSymbolFont{boldletters}{OT1}{cmr}{bx}{n}
\DeclareSymbolFontAlphabet{\mathcal}{symbols}
\DeclareMathSymbol{x}{\mathalpha}{boldletters}{`x}
\DeclareMathSymbol{z}{\mathalpha}{boldletters}{`z}
\DeclareMathSymbol{c}{\mathalpha}{boldletters}{`c}
\DeclareMathSymbol{o}{\mathalpha}{boldletters}{`o}
\DeclareMathSymbol{R}{\mathalpha}{symbols}{`R}
\DeclareMathSymbol{F}{\mathalpha}{symbols}{`F}
\DeclareMathSymbol{G}{\mathalpha}{symbols}{`G}

\DeclareMathOperator{\EX}{\mathbb{E}}

\newcommand{\recog}{{\mathchoice{}{}{\scriptscriptstyle}{}R}}
\newcommand{\ed}{{\mathchoice{}{}{\scriptscriptstyle}{} F}}
\newcommand{\gen}{{\mathchoice{}{}{\scriptscriptstyle}{} G}}
\DeclareMathOperator*{\argmax}{argmax}

\title{Unsupervised Discrete Sentence Representation Learning for Interpretable Neural Dialog Generation}

\author{Tiancheng Zhao, Kyusong Lee and Maxine Eskenazi \\
 Language Technologies Institute \\
  Carnegie Mellon University \\
  Pittsburgh, Pennsylvania, USA \\
  {\tt \{tianchez, kyusongl, max+\}@cs.cmu.edu} \\}

\date{}

\begin{document}
\maketitle
\begin{abstract}
  The encoder-decoder dialog model is one of the most prominent methods used to build dialog systems in complex domains. Yet it is limited because it cannot output interpretable actions as in traditional systems, which hinders humans from understanding its generation process. We present an unsupervised discrete sentence representation learning method that can integrate with any existing encoder-decoder dialog models for interpretable response generation. Building upon variational autoencoders (VAEs), we present two novel models, DI-VAE and DI-VST that improve VAEs and can discover interpretable semantics via either auto encoding or context predicting. Our methods have been validated on real-world dialog datasets to discover semantic representations and enhance encoder-decoder models with interpretable generation.\footnote{Data and code are available at \url{https://github.com/snakeztc/NeuralDialog-LAED}.}
\end{abstract}

\section{Introduction}
Classic dialog systems rely on developing a meaning representation to represent the utterances from both the machine and human users~\cite{larsson2000information,bohus2007olympus}. The dialog manager of a conventional dialog system outputs the system's next action in a semantic frame that usually contains hand-crafted dialog acts and slot values~\cite{williams2007partially}. Then a natural language generation module is used to generate the system's output in natural language based on the given semantic frame. This approach suffers from generalization to more complex domains because it soon become intractable to manually design a frame representation that covers all of the fine-grained system actions. The recently developed neural dialog system is one of the most prominent frameworks for developing dialog agents in complex domains. The basic model is based on encoder-decoder networks~\cite{cho2014learning} and can learn to generate system responses without the need for hand-crafted meaning representations and other annotations. 
\begin{figure}[ht]
    \centering
    \includegraphics[width=0.46\textwidth]{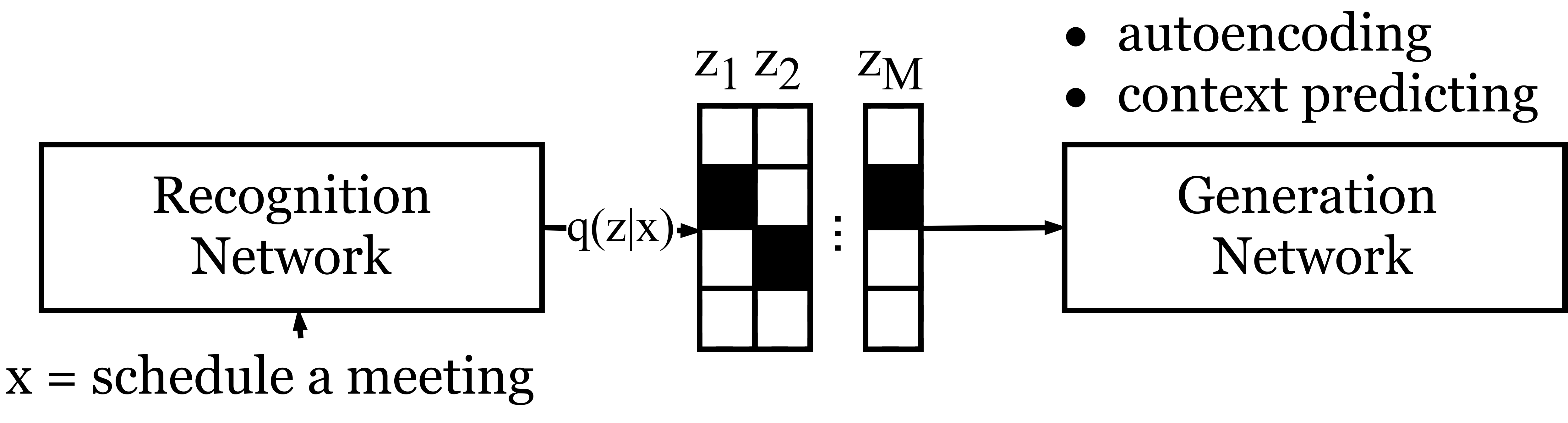}
    \caption{Our proposed models learn a set of discrete variables to represent sentences by either autoencoding or context prediction.}
    \label{fig:goal}
\end{figure}

Although generative dialog models have advanced rapidly~\cite{serban2016hierarchical,li2016persona,zhao2017learning}, they cannot provide interpretable system actions as in the conventional dialog systems. This inability limits the effectiveness of generative dialog models in several ways. First, having interpretable system actions enables human to understand the behavior of a dialog system and better interpret the system intentions. Also, modeling the high-level decision-making policy in dialogs enables useful generalization and data-efficient domain adaptation~\cite{gavsic2010gaussian}. Therefore, the motivation of this paper is to develop an unsupervised neural recognition model that can discover interpretable meaning representations of utterances (denoted as \textit{latent actions}) as a set of discrete latent variables from a large unlabelled corpus as shown in Figure~\ref{fig:goal}. The discovered meaning representations will then be integrated with encoder decoder networks to achieve interpretable dialog generation while preserving all the merit of neural dialog systems. 

We focus on learning discrete latent representations instead of dense continuous ones because discrete variables are easier to interpret~\cite{van2017neural} and can naturally correspond to categories in natural languages, e.g. topics, dialog acts and etc. Despite the difficulty of learning discrete latent variables in neural networks, the recently proposed Gumbel-Softmax offers a reliable way to back-propagate through discrete variables~\cite{maddison2016concrete,jang2016categorical}. However, we found a simple combination of sentence variational autoencoders (VAEs)~\cite{bowman2015generating} and Gumbel-Softmax fails to learn meaningful discrete representations. We then highlight the anti-information limitation of the evidence lowerbound objective (ELBO) in VAEs and improve it by proposing Discrete Information VAE (DI-VAE) that maximizes the mutual information between data and latent actions. We further enrich the learning signals beyond auto encoding by extending Skip Thought~\cite{kiros2015skip} to Discrete Information Variational Skip Thought (DI-VST) that learns sentence-level distributional semantics. Finally, an integration mechanism is presented that combines the learned latent actions with encoder decoder models. 

The proposed systems are tested on several real-world dialog datasets. Experiments show that the proposed methods significantly outperform the standard VAEs and can discover meaningful latent actions from these datasets. Also, experiments confirm the effectiveness of the proposed integration mechanism and show that the learned latent actions can control the sentence-level attributes of the generated responses and provide human-interpretable meaning representations. 


\section{Related Work}
Our work is closely related to research in latent variable dialog models. The majority of models are based on Conditional Variational Autoencoders (CVAEs)~\cite{serban2016hierarchical,cao2017latent} with continuous latent variables to better model the response distribution and encourage diverse responses. Zhao et al.,~\shortcite{zhao2017learning} further introduced dialog acts to guide the learning of the CVAEs. Discrete latent variables have also been used for task-oriented dialog systems~\cite{wen2017latent}, where the latent space is used to represent intention. The second line of related work is enriching the dialog context encoder with more fine-grained information than the dialog history. Li et al.,~\shortcite{li2016persona} captured speakers' characteristics by encoding background information and speaking style into the distributed embeddings. Xing et al.,~\shortcite{xing2016topic} maintain topic encoding based on Latent Dirichlet Allocation (LDA)~\cite{blei2003latent} of the conversation to encourage the model to output more topic coherent responses. 

The proposed method also relates to sentence representation learning using neural networks. Most work learns continuous distributed representations of sentences from various learning signals~\cite{hill2016learning}, e.g. the Skip Thought learns representations by predicting the previous and next sentences~\cite{kiros2015skip}. Another area of work focused on learning regularized continuous sentence representation, which enables sentence generation by sampling the latent space~\cite{bowman2015generating,kim2017adversarially}. There is less work on discrete sentence representations due to the difficulty of passing gradients through discrete outputs. The recently developed Gumbel Softmax~\cite{jang2016categorical,maddison2016concrete} and vector quantization~\cite{van2017neural} enable us to train discrete variables. Notably, discrete variable models have been proposed to discover document topics~\cite{miao2016neural} and semi-supervised sequence transaction~\cite{zhou2017multi}

Our work differs from these as follows: (1) we focus on learning interpretable variables; in prior research the semantics of latent variables are mostly ignored in the dialog generation setting. (2) we improve the learning objective for discrete VAEs and overcome the well-known posterior collapsing issue~\cite{bowman2015generating,chen2016variational}. (3) we focus on unsupervised learning of salient features in dialog responses instead of hand-crafted features.

\section{Proposed Methods}
Our formulation contains three random variables: the dialog context $c$, the response $x$ and the latent action $z$. The context often contains the discourse history in the format of a list of utterances. The response is an utterance that contains a list of word tokens. The latent action is a set of discrete variables that define high-level attributes of $x$. Before introducing the proposed framework, we first identify two key properties that are essential in order for $z$ to be \textit{interpretable}:
\begin{enumerate}[itemsep=1mm]
    \item $z$ should capture salient sentence-level features about the response $x$. 
    \item The meaning of latent symbols $z$ should be independent of the context $c$. 
\end{enumerate}
The first property is self-evident. The second can be explained: assume $z$ contains a single discrete variable with $K$ classes. Since the context $c$ can be any dialog history, if the meaning of each class changes given a different context, then it is difficult to extract an intuitive interpretation by only looking at all responses with class $k \in [1, K]$. Therefore, the second property looks for latent actions that have \textit{context-independent semantics} so that each assignment of $z$ conveys the same meaning in all dialog contexts. 

With the above definition of interpretable latent actions, we first introduce a recognition network $R:q_{\recog}(z|x)$ and a generation network $G$. The role of $R$ is to map an sentence to the latent variable $z$ and the generator $G$ defines the learning signals that will be used to train $z$'s representation. Notably, our recognition network $R$ does not depend on the context $c$ as has been the case in prior work~\cite{serban2016hierarchical}. The motivation of this design is to encourage $z$ to capture context-independent semantics, which are further elaborated in Section~\ref{sec:why_no_cvae}. With the $z$ learned by $R$ and $G$, we then introduce an encoder decoder network $F: p_{\ed}(x|z,c)$ and and a policy network $\pi: p_\pi(z|c)$. At test time, given a context $c$, the policy network and encoder decoder will work together to generate the next response via $\Tilde{x}= p_{\ed}(x|z \sim p_\pi(z|c),c)$. In short, $R$, $G$, $F$ and $\pi$ are the four components that comprise our proposed framework. The next section will first focus on developing $R$ and $G$ for learning interpretable $z$ and then will move on to integrating $R$ with $F$ and $\pi$ in Section~\ref{sec:integration}. 

\subsection{Learning Sentence Representations from Auto-Encoding}
Our baseline model is a sentence VAE with discrete latent space. We use an RNN as the recognition network to encode the response $x$. Its last hidden state $h^R_{|x|}$ is used to represent $x$. We define $z$ to be a set of K-way categorical variables $z=\{z_1...z_m...z_M\}$, where $M$ is the number of variables. For each $z_m$, its posterior distribution is defined as $q_{\recog}(z_m|x)=\text{Softmax}(W_q h^R_{|x|}+b_q)$. During training, we use the Gumbel-Softmax trick to sample from this distribution and obtain low-variance gradients. To map the latent samples to the initial state of the decoder RNN, we define $\{e_1...e_m...e_M\}$ where $e_m \in \mathbb{R}^{K \times D}$ and $D$ is the generator cell size. Thus the initial state of the generator is: $h^G_0=\sum_{m=1}^M e_m(z_m)$. Finally, the generator RNN is used to reconstruct the response given $h^G_0$. VAEs is trained to maxmimize the evidence lowerbound objective (ELBO)~\cite{kingma2013auto}. For simplicity, later discussion drops the subscript $m$ in $z_m$ and assumes a single latent $z$. Since each $z_m$ is independent, we can easily extend the results below to multiple variables.

\subsubsection{Anti-Information Limitation of ELBO}
It is well-known that sentence VAEs are hard to train because of the posterior collapse issue. Many empirical solutions have been proposed: weakening the decoder, adding auxiliary loss etc.~\cite{bowman2015generating,chen2016variational,zhao2017learning}. We argue that the posterior collapse issue lies in ELBO and we offer a novel decomposition to understand its behavior. First, instead of writing ELBO for a single data point, we write it as an expectation over a dataset:
\begin{align}
\begin{split}
   {\mathcal{L}_\text{VAE}}  = \EX_{x} &[\EX_{q_{\recog}(z|x)} [\log p_{\gen}(x|z)] \\
   &- \text{KL}(q_{\recog}(z|x)\|p(z))]
\end{split}
\end{align}
We can expand the KL term as Eq.~\ref{eq:kl_decompose} (derivations in Appendix~\ref{sec:anti-info-detail}) and rewrite ELBO as:
\begin{align}
    \label{eq:kl_decompose}
    \EX_x [\text{KL}(q_{\recog}(z|x)\|p(z))] & = \\ \nonumber
         I(Z, X) + & \text{KL}(q(z)\|p(z)) 
\end{align}
\begin{align}
    \begin{split}
    {\mathcal{L}_\text{VAE}} = &\EX_{q(z|x)p(x)} [\log p(x|z)]\\ 
                               &- I(Z, X) - \text{KL}(q(z)\|p(z))
    \end{split}
    \label{eq:elbo_anti_mi}
\end{align}
where $q(z)=\EX_x[q_{\recog}(z|x)]$ and $I(Z,X)$ is the mutual information between $Z$ and $X$. This expansion shows that the KL term in ELBO is trying to reduce the mutual information between latent variables and the input data, which explains why VAEs often ignore the latent variable, especially when equipped with powerful decoders.

\subsubsection{VAE with Information Maximization and Batch Prior Regularization}
A natural solution to correct the anti-information issue in Eq.~\ref{eq:elbo_anti_mi} is to maximize both the data likelihood lowerbound and the mutual information between $z$ and the input data:
\begin{align}
    \begin{split}
    & \mathcal{L}_\text{VAE} + I(Z,X) = \\ 
    &\EX_{q_{\recog}(z|x)p(x)} [\log p_{\gen}(x|z)] - \text{KL}(q(z)\|p(z))
    \end{split}
    \label{eq:elbo_mi}
\end{align}

Therefore, jointly optimizing ELBO and mutual information simply cancels out the information-discouraging term. Also, we can still sample from the prior distribution for generation because of $\text{KL}(q(z)\|p(z))$. Eq.~\ref{eq:elbo_mi} is similar to the objectives used in adversarial autoencoders~\cite{makhzani2015adversarial,kim2017adversarially}. Our derivation provides a theoretical justification to their superior performance. Notably, Eq.~\ref{eq:elbo_mi} arrives at the same loss function proposed in infoVAE~\cite{zhao2017infovae}. However, our derivation is different, offering a new way to understand ELBO behavior.

The remaining challenge is how to minimize $\text{KL}(q(z)\|p(z))$, since $q(z)$ is an expectation over $q(z|x)$. When $z$ is continuous, prior work has used adversarial training~\cite{makhzani2015adversarial,kim2017adversarially} or Maximum Mean Discrepancy (MMD)~\cite{zhao2017infovae} to regularize $q(z)$. It turns out that minimizing $\text{KL}(q(z)\|p(z))$ for discrete $z$ is much simpler than its continuous counterparts. Let $x_n$ be a sample from a batch of $N$ data points. Then we have:
\begin{align}
    q(z) \approx \frac{1}{N} \sum_{n=1}^{N} q(z|x_n) = q'(z)
\end{align}
where $q'(z)$ is a mixture of softmax from the posteriors $q(z|x_n)$ of each $x_n$. We can approximate $\text{KL}(q(z)\|p(z))$ by:
\begin{align}
    \text{KL}(q'(z)\|p(z)) = \sum_{k=1}^{K} q'(z=k) \log\frac{q'(z=k)}{p(z=k)}
    \label{eq:qz}
\end{align}
We refer to Eq.~\ref{eq:qz} as Batch Prior Regularization (BPR). When $N$ approaches infinity, $q'(z)$ approaches the true marginal distribution of $q(z)$. In practice, we only need to use the data from each mini-batch assuming that the mini batches are randomized. Last, BPR is fundamentally different from multiplying a coefficient $<1$ to anneal the KL term in VAE~\cite{bowman2015generating}. This is because BPR is a non-linear operation \textit{log\_sum\_exp}. For later discussion, we denote our discrete infoVAE with BPR as DI-VAE.

\subsection{Learning Sentence Representations from the Context}
DI-VAE infers sentence representations by reconstruction of the input sentence. Past research in distributional semantics has suggested the meaning of language can be inferred from the adjacent context~\cite{harris1954distributional,hill2016learning}. The distributional hypothesis is especially applicable to dialog since the utterance meaning is highly contextual. For example, the dialog act is a well-known utterance feature and depends on dialog state~\cite{austin1975things,stolcke2000dialogue}. Thus, we introduce a second type of latent action based on sentence-level distributional semantics. 

Skip thought (ST) is a powerful sentence representation that captures contextual information~\cite{kiros2015skip}. ST uses an RNN to encode a sentence, and then uses the resulting sentence representation to predict the previous and next sentences. Inspired by ST's robust performance across multiple tasks~\cite{hill2016learning}, we adapt our DI-VAE to Discrete Information Variational Skip Thought (DI-VST) to learn discrete latent actions that model distributional semantics of sentences. We use the same recognition network from DI-VAE to output $z$'s posterior distribution $q_{\recog}(z|x)$. Given the samples from $q_{\recog}(z|x)$, two RNN generators are used to predict the previous sentence $x_p$ and the next sentences $x_n$. Finally, the learning objective is to maximize:
\begin{align}
    \begin{split}
    \mathcal{L}_\text{DI-VST} &= \EX_{q_{\recog}(z|x)p(x))} [\log(p^n_{\gen}(x_n|z) p^p_{\gen}(x_p|z))]\\ 
    & - \text{KL}(q(z)\|p(z))
    \end{split}
    \label{eq:info_vst}
\end{align}

\subsection{Integration with Encoder Decoders}
\label{sec:integration}
We now describe how to integrate a given $q_{\recog}(z|x)$ with an encoder decoder and a policy network. Let the dialog context $c$ be a sequence of utterances. Then a dialog context encoder network can encode the dialog context into a distributed representation $h^e=F^e(c)$. The decoder $F^d$ can generate the responses $\Tilde{x}=F^d(h^e, z)$ using samples from $q_{\recog}(z|x)$. Meanwhile, we train $\pi$ to predict the aggregated posterior $\EX_{p(x|c)}[q_{\recog}(z|x)]$ from $c$ via maximum likelihood training. This model is referred as Latent Action Encoder Decoder (LAED) with the following objective.
\begin{align}
    \begin{split}
            \mathcal{L}_\text{LAED}(\theta_F, \theta_{\pi}) =&  \\
                            \EX_{q_{\recog}(z|x)p(x,c)} [ \log & p_{\pi}(z|c) + \log p_{\ed}(x|z, c)] 
    \end{split}
    \label{eq:la-ndm}
\end{align}
Also simply augmenting the inputs of the decoders with latent action does not guarantee that the generated response exhibits the attributes of the give action. Thus we use the controllable text generation framework~\cite{hu2017toward} by introducing $\mathcal{L}_\text{Attr}$, which reuses the same recognition network $q_{\recog}(z|x)$ as a fixed discriminator to penalize the decoder if its generated responses do not reflect the attributes in $z$. 
\begin{equation}
    \mathcal{L}_\text{Attr}(\theta_{\ed}) = \EX_{q_{\recog}(z|x)p(c,x)}[\log q_{\recog}(z|\mathcal{F}(c, z))]
    \label{eq:attr_loss}
\end{equation}
Since it is not possible to propagate gradients through the discrete outputs at $F^d$ at each word step, we use a deterministic continuous relaxation~\cite{hu2017toward} by replacing output of $F^d$ with the probability of each word. Let $o_t$ be the normalized probability at step $t\in [1, |x|]$, the inputs to $q_{\recog}$ at time $t$ are then the sum of word embeddings weighted by $o_t$, i.e. $h^R_{t}=\text{RNN}(h^R_{t-1}, \mathbf{E}o_t)$ and $\mathbf{E}$ is the word embedding matrix. Finally  this loss is combined with $\mathcal{L}_\text{LAED}$  and a hyperparameter $\lambda$ to have Attribute Forcing LAED. 
\begin{equation}
    \mathcal{L}_\text{attrLAED} = \mathcal{L}_\text{LAED} + \lambda \mathcal{L}_\text{Attr}
\end{equation}

\subsection{Relationship with Conditional VAEs}
\label{sec:why_no_cvae}
It is not hard to see $\mathcal{L}_\text{LAED}$ is closely related to the objective of CVAEs for dialog generation~\cite{serban2016hierarchical,zhao2017learning}, which is:
\begin{equation}
    \mathcal{L_\text{CVAE}}= \EX_q[\log p(x|z,c)]-\text{KL}(q(z|x,c)\| p(z|c))
\end{equation}
Despite their similarities, we highlight the key differences that prohibit CVAE from achieving interpretable dialog generation. First $\mathcal{L}_\text{CVAE}$ encourages $I(x, z|c)$~\cite{agakov2005variational}, which learns $z$ that capture context-dependent semantics. More intuitively, $z$ in CVAE is trained to generate $x$ via $p(x|z,c)$ so the meaning of learned $z$ can only be interpreted along with its context $c$. Therefore this violates our goal of learning context-independent semantics. Our methods learn $q_{\recog}(z|x)$ that only depends on $x$ and trains $q_{\recog}$ separately to ensure the semantics of $z$ are interpretable standalone. 

\section{Experiments and Results}
\label{sec:exp_setup}
The proposed methods are evaluated on four datasets. The first corpus is Penn Treebank (PTB)~\cite{marcus1993building} used to evaluate sentence VAEs~\cite{bowman2015generating}. We used the version pre-processed by Mikolov~\cite{mikolov2010recurrent}. The second dataset is the Stanford Multi-Domain Dialog (SMD) dataset that contains 3,031 human-Woz, task-oriented dialogs collected from 3 different domains (navigation, weather and scheduling)~\cite{eric2017key}. The other two datasets are chat-oriented data: Daily Dialog (DD) and Switchboard (SW)~\cite{godfrey1997switchboard}, which are used to test whether our methods can generalize beyond task-oriented dialogs but also to to open-domain chatting. DD contains 13,118 multi-turn human-human dialogs annotated with dialog acts and emotions.~\cite{li2017dailydialog}. SW has 2,400 human-human telephone conversations that are annotated with topics and dialog acts. SW is a more challenging dataset because it is transcribed from speech which contains complex spoken language phenomenon, e.g. hesitation, self-repair etc.

\subsection{Comparing Discrete Sentence Representation Models}
The first experiment used PTB and DD to evaluate the performance of the proposed methods in learning discrete sentence representations. We implemented DI-VAE and DI-VST using GRU-RNN~\cite{chung2014empirical} and trained them using Adam~\cite{kingma2014adam}. Besides the proposed methods, the following baselines are compared. \textbf{Unregularized models}: removing the KL$(q|p)$ term from DI-VAE and DI-VST leads to a simple discrete autoencoder (DAE) and discrete skip thought (DST) with stochastic discrete hidden units. \textbf{ELBO models}: the basic discrete sentence VAE (DVAE) or variational skip thought (DVST) that optimizes ELBO with regularization term KL$(q(z|x)\|p(z))$. We found that standard training failed to learn informative latent actions for either DVAE or DVST because of the posterior collapse. Therefore, KL-annealing~\cite{bowman2015generating} and bag-of-word loss~\cite{zhao2017learning} are used to force these two models learn meaningful representations. We also include the results for VAE with continuous latent variables reported on the same PTB~\cite{zhao2017learning}. Additionally, we report the perplexity from a standard GRU-RNN language model~\cite{zaremba2014recurrent}. 

The evaluation metrics include reconstruction perplexity (PPL), KL$(q(z)\|p(z))$ and the mutual information between input data and latent variables $I(x,z)$. Intuitively a good model should achieve low perplexity and KL distance, and simultaneously achieve high  $I(x,z)$. The discrete latent space for all models are $M$=20 and $K$=10. Mini-batch size is 30.
\begin{table}[ht]
\centering
\begin{tabular}{p{0.04\textwidth}|p{0.08\textwidth}|p{0.09\textwidth}p{0.07\textwidth}p{0.07\textwidth}}\hline
Dom    & Model        & PPL              & $\text{KL}(q\|p)$  & $I(x,z)$        \\ \hline
PTB    & RNNLM        & 116.22           & -                  & -               \\
       & VAE         & 73.49            & 15.94*             & -               \\
       & DAE          & 66.49            & 2.20               & 0.349           \\
       & DVAE        & 70.84            & 0.315              & 0.286           \\ 
       & DI-VAE       & \textbf{52.53}   & \textbf{0.133}     & \textbf{1.18}   \\ \hline
DD     & RNNLM        & 31.15                   & -                 & -             \\
       & DST          & $x_p$:28.23 $x_n$:28.16        & 0.588             & \textbf{1.359}         \\ 
       & DVST        & $x_p$:30.36  $x_n$:30.71        & \textbf{0.007}             & 0.081         \\ 
       & DI-VST       & $x_p$:\textbf{28.04}  $x_n$:\textbf{27.94}        & 0.088             & 1.028         \\ \hline 
\end{tabular}
\caption{Results for various discrete sentence representations. The KL for VAE is KL$(q(z|x)\|p(z))$ instead of KL$(q(z)\|p(z))$~\cite{zhao2017learning}}
\label{tbl:vae-main}
\end{table}

Table~\ref{tbl:vae-main} shows that all models achieve better perplexity than an RNNLM, which shows they manage to learn meaningful $q(z|x)$. First, for auto-encoding models, DI-VAE is able to achieve the best results in all metrics compared other methods. We found DAEs quickly learn to reconstruct the input but they are prone to overfitting during training, which leads to lower performance on the test data compared to DI-VAE. Also, since there is no regularization term in the latent space, $q(z)$ is very different from the $p(z)$ which prohibits us from generating sentences from the latent space. In fact, DI-VAE enjoys the same linear interpolation properties reported in~\cite{bowman2015generating} (See Appendix~\ref{sec:vae-interpolation}). As for DVAEs, it achieves zero $I(x,z)$ in standard training and only manages to learn some information when training with KL-annealing and bag-of-word loss. On the other hand, our methods achieve robust performance without the need for additional processing. Similarly, the proposed DI-VST is able to achieve the lowest PPL and similar KL compared to the strongly regularized DVST. Interestingly, although DST is able to achieve the highest $I(x,z)$, but PPL is not further improved. These results confirm the effectiveness of the proposed BPR in terms of regularizing $q(z)$ while learning meaningful posterior $q(z|x)$.  

In order to understand BPR's sensitivity to batch size $N$, a follow-up experiment varied the batch size from 2 to 60 (If $N$=1, DI-VAE is equivalent to DVAE). 
\begin{figure}[ht]
    \centering
    \includegraphics[width=0.45\textwidth]{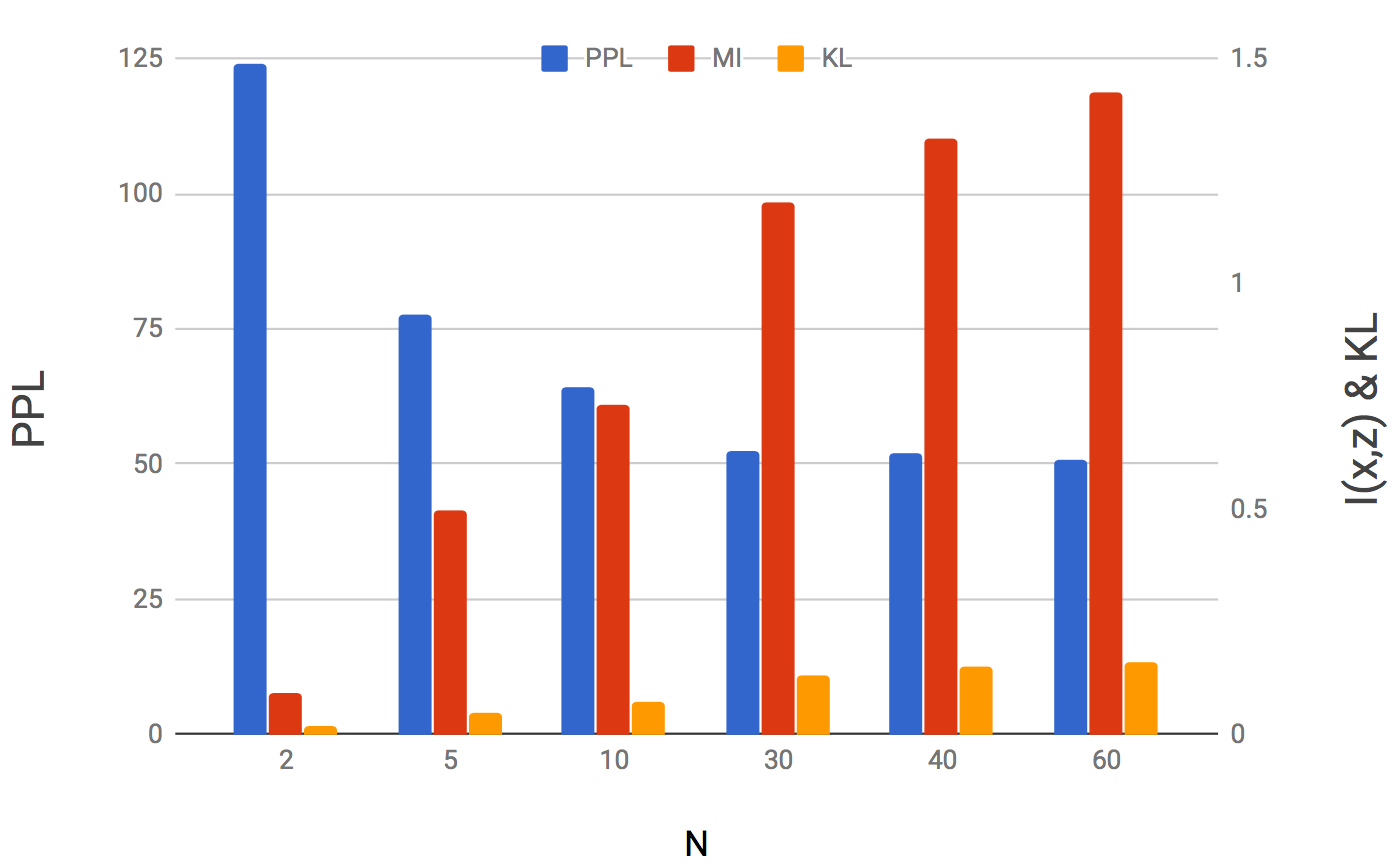}
    \caption{Perplexity and $I(x,z)$ on PTB by varying batch size $N$. BPR works better for larger $N$.}
    \label{fig:bpr}
\end{figure}
Figure~\ref{fig:bpr} show that as $N$ increases, perplexity, $I(x,z)$ monotonically improves, while $\text{KL}(q\|p)$ only increases from 0 to 0.159. After $N>30$, the performance plateaus. Therefore, using mini-batch is an efficient trade-off between $q(z)$ estimation and computation speed.

The last experiment in this section investigates the relation between representation learning and the dimension of the latent space. We set a fixed budget by restricting the maximum number of modes to be about 1000, i.e. ${K}^{M} \approx 1000$. We then vary the latent space size and report the same evaluation metrics. Table~\ref{tbl:vae-size} shows that models with multiple small latent variables perform significantly better than those with large and few latent variables. 
\begin{table}[ht]
\centering
\begin{tabular}{llllll}\hline
K, M     & $K^M$       & PPL       & $\text{KL}(q\|p)$ & $I(x,z)$ \\ \hline
1000, 1  & 1000        & 75.61     & 0.032             & 0.335     \\
10, 3    & 1000        & 71.42     & 0.071             & 0.607     \\
4, 5     & 1024        & 68.43     & 0.088             & 0.809     \\ \hline
\end{tabular}
\caption{DI-VAE on PTB with different latent dimensions under the same budget.}
\label{tbl:vae-size}
\end{table}

\subsection{Interpreting Latent Actions}
The next question is to interpret the meaning of the learned latent action symbols. To achieve this, the latent action of an utterance $x_n$ is obtained from a greedy mapping: $a_n=\argmax_k q_{\recog}(z=k|x_n)$. We set $M$=3 and $K$=5, so that there are at most 125 different latent actions, and each $x_n$ can now be represented by $a_1$-$a_2$-$a_3$, e.g. ``How are you?'' $\rightarrow$ 1-4-2. Assuming that we have access to manually clustered data according to certain classes (e.g. dialog acts), it is unfair to use classic cluster measures~\cite{vinh2010information} to evaluate the clusters from latent actions. This is because the uniform prior $p(z)$ evenly distribute the data to all possible latent actions, so that it is expected that frequent classes will be assigned to several latent actions. Thus we utilize the \textit{homogeneity} metric~\cite{rosenberg2007v} that measures if each latent action contains only members of a single class. We tested this on the SW and DD, which contain human annotated features and we report the latent actions' homogeneity w.r.t these features in Table~\ref{tbl:auto-cluster}. 
\begin{table}[ht]
\centering
\begin{tabular}{l|ll|ll}\hline
                 & \textbf{SW}  &           & \textbf{DD}    &           \\ \hline
                 & Act          & Topic     & Act            & Emotion    \\
DI-VAE           & 0.48         & 0.08      & 0.18           & 0.09        \\
DI-VST           & 0.33         & 0.13      & 0.34           & 0.12         \\ \hline
\end{tabular}
\caption{Homogeneity results (bounded [0, 1]).}
\label{tbl:auto-cluster}
\end{table}
On DD, results show DI-VST works better than DI-VAE in terms of creating actions that are more coherent for emotion and dialog acts. The results are interesting on SW since DI-VST performs worse on dialog acts than DI-VAE. One reason is that the dialog acts in SW are more fine-grained (42 acts) than the ones in DD (5 acts) so that distinguishing utterances based on words in $x$ is more important than the information in the neighbouring utterances.

We then apply the proposed methods to SMD which has no manual annotation and contains task-oriented dialogs. Two experts are shown 5 randomly selected utterances from each latent action and are asked to give an action name that can describe as many of the utterances as possible. Then an Amazon Mechanical Turk study is conducted to evaluate whether other utterances from the same latent action match these titles. 5 workers see the action name and a different group of 5 utterances from that latent action. They are asked to select all utterances that belong to the given actions, which tests the homogeneity of the utterances falling in the same cluster. Negative samples are included to prevent random selection. Table~\ref{tbl:amt} shows that both methods work well and DI-VST achieved better homogeneity than DI-VAE. 
\begin{table}[ht]
\centering
\begin{tabular}{l|llll}\hline
Model            & Exp Agree              & Worker $\kappa$      & Match Rate   \\ \hline
DI-VAE           & 85.6\%                 & 0.52                 & 71.3\%                           \\
DI-VST           & 93.3\%                 & 0.48                 & 74.9\%           \\ \hline
\end{tabular}
\caption{Human evaluation results on judging the homogeneity of latent actions in SMD.}
\label{tbl:amt}
\end{table}

Since DI-VAE is trained to reconstruct its input and DI-VST is trained to model the context, they group utterances in different ways. For example, DI-VST would group ``Can I get a restaurant'', ``I am looking for a restaurant'' into one action where DI-VAE may denote two actions for them. Finally, Table~\ref{tbl:code-meaning} shows sample annotation results, which show cases of the different types of latent actions discovered by our models. 
\begin{table}[ht]
\centering
\small
\begin{tabular}{p{0.07\textwidth}|p{0.09\textwidth}|p{0.24\textwidth}}\hline
\textbf{Model}    & \textbf{Action}         & \textbf{Sample utterance}   \\ \hline
DI-VAE       & scheduling                   & - sys: okay, scheduling a yoga activity with Tom for the 8th at 2pm.                \\
             &                              & - sys: okay, scheduling a meeting for 6 pm on Tuesday with your boss to go over the quarterly report. \\ \cline{2-3}
             & requests                     & - usr: find out if it 's supposed to rain \\ 
             &                              & - usr: find nearest coffee shop \\ \hline  
DI-VST       & ask schedule info            & - usr: when is my football activity and who is going with me?                \\
             &                              & - usr: tell me when my dentist appointment is? \\ \cline{2-3}
             & requests                     & - usr: how about other coffee? \\ 
             &                              & - usr: 11 am please \\ \hline           
\end{tabular}
\label{tbl:code-meaning}
\caption{Example latent actions discovered in SMD using our methods.}
\end{table}
 
\subsection{Dialog Response Generation with Latent Actions}
Finally we implement an LAED as follows. The encoder is a hierarchical recurrent encoder~\cite{serban2016hierarchical} with bi-directional GRU-RNNs as the utterance encoder and a second GRU-RNN as the discourse encoder. The discourse encoder output its last hidden state $h^e_{|x|}$. The decoder is another GRU-RNN and its initial state of the decoder is obtained by $h^d_0=h^e_{|x|} + \sum_{m=1}^M e_m(z_m)$, where $z$ comes from the recognition network of the proposed methods. The policy network $\pi$ is a 2-layer multi-layer perceptron (MLP) that models $p_{\pi}(z|h^e_{|x|})$. We use up to the previous 10 utterances as the dialog context and denote the LAED using DI-VAE latent actions as AE-ED and the one uses DI-VST as ST-ED.  

First we need to confirm whether an LAED can generate responses that are consistent with the semantics of a given $z$. To answer this, we use a pre-trained recognition network $R$ to check if a generated response carries the attributes in the given action. We generate dialog responses on a test dataset via $\Tilde{x}=F(z \sim \pi(c), c)$ with greedy RNN decoding. The generated responses are passed into the $R$ and we measure \textit{attribute accuracy} by counting $\Tilde{x}$ as correct if $z=\argmax_k q_{\recog}(k|\Tilde{x})$. 
\begin{table}[ht]
\centering
\label{tbl:adv-eval}
\begin{tabular}{l|ll|ll}\hline
Domain     & AE-ED     & +$L_\text{attr}$  & ST-ED       & +$L_\text{attr}$    \\ \hline
SMD       & 93.5\%    & 94.8\%     & 91.9\%      & 93.8\%       \\ 
DD        & 88.4\%    & 93.6\%     & 78.5\%      & 86.1\%       \\ 
SW        & 84.7\%    & 94.6\%     & 57.3\%      & 61.3\%       \\ \hline
\end{tabular}
\caption{Results for attribute accuracy with and without attribute loss.}
\end{table}
Table~\ref{tbl:adv-eval} shows our generated responses are highly consistent with the given latent actions. Also, latent actions from DI-VAE achieve higher attribute accuracy than the ones from DI-VST, because $z$ from auto-encoding is explicitly trained for $x$ reconstruction. Adding $\mathcal{L}_{attr}$ is effective in forcing the decoder to take $z$ into account during its generation, which helps the most in more challenging open-domain chatting data, e.g. SW and DD. The accuracy of ST-ED on SW is worse than the other two datasets. The reason is that SW contains many short utterances that can be either a continuation of the same speaker or a new turn from the other speaker, whereas the responses in the other two domains are always followed by a different speaker. The more complex context pattern in SW may require special treatment. We leave it for future work. 

The second experiment checks if the policy network $\pi$ is able to predict the right latent action given just the dialog context. We report both accuracy, i.e. $\argmax_k q_{\recog}(k|x)=\argmax_{k'} p_{\pi}(k'|c)$ and perplexity of $p_{\pi}(z|c)$. The perplexity measure is more useful for open domain dialogs because decision-making in complex dialogs is often one-to-many given a similar context~\cite{zhao2017learning}.
\begin{table}[ht]
\centering
\label{tbl:dialog-main}
\begin{tabular}{l|ll}\hline
              & \multicolumn{2}{l}{SMD}            \\ \hline
AE-ED         & \multicolumn{2}{l}{3.045 (51.5\% sys 52.4\% usr 50.5\%)}         \\ 
ST-ED         & \multicolumn{2}{l}{1.695 (75.5\% sys 82.1\% usr 69.2\%)}          \\ \hline
              & DD                  & SW                                \\ \hline
AE-ED         & 4.47 (35.8\%)       & 4.46 (31.68\%)   \\
ST-ED         & 3.89 (47.5\%)       & 3.68 (33.2\%)  \\\hline
\end{tabular}
\caption{Performance of policy network. $\mathcal{L}_{attr}$ is included in training.}
\label{tbl:policy}
\end{table}
Table~\ref{tbl:policy} shows the prediction scores on the three dialog datasets. These scores provide useful insights to understand the complexity of a dialog dataset. For example, accuracy on open-domain chatting is harder than the task-oriented SMD data. Also, it is intuitive that predicting system actions is easier than predicting user actions on SMD. Also, in general the prediction scores for ST-ED are higher the ones for AE-ED. The reason is related to our previous discussion about the granularity of the latent actions. Since latent actions from DI-VST mainly model the the type of utterances used in certain types of context, it is easier for the policy network to predict latent actions from DI-VST. Therefore, choosing the type of latent actions is a design choice and depends on the type of interpretability that is needed. We finish with an example generated from the two variants of LAED on SMD as shown in Table~\ref{tbl:dialog-example}. Given a dialog context, our systems are able to output a probability distribution over different latent actions that have interpretable meaning along with their natural language realizations. 
\begin{table}[ht]
\centering
\small
\label{my-label}
\begin{tabular}{p{0.05\textwidth}|p{0.11\textwidth}p{0.25\textwidth}} \hline
$c$     & \multicolumn{2}{p{0.35\textwidth}}{\textbf{usr}: Where does my friend live?} \\ \hline
Model       & Action                    & Generated Responses                         \\ \hline
AE-ED       & give loc info                      & - Tom is at 753 University Ave, and a road block.     \\
            & $p(z|c)$=0.34                     & - Comfort Inn is at 7 miles away.                     \\  \cline{2-3}
            & give user info                  & - Your home address is 5671 barringer street.         \\
            & $p(z|c)$=0.22                          & - Your home is at 10 ames street.                     \\\hline
ST-ED       & give loc info        & - Jill's house is 8 miles away at 347 Alta Mesa Ave. \\ 
            & $p(z|c)$=0.93                    & - Jill lives at 347 Alta Mesa Ave.                \\\hline
\end{tabular}
\caption{Interpretable dialog generation on SMD with top probable latent actions. AE-ED predicts more fine-grained but more error-prone actions.}
\label{tbl:dialog-example}
\end{table}

\section{Conclusion and Future Work}
This paper presents a novel unsupervised framework that enables the discovery of discrete latent actions and interpretable dialog response generation. Our main contributions reside in the two sentence representation models DI-VAE and DI-VST, and their integration with the encoder decoder models. Experiments show the proposed methods outperform strong baselines in learning discrete latent variables and showcase the effectiveness of interpretable dialog response generation. Our findings also suggest promising future research directions, including learning better context-based latent actions and using reinforcement learning to adapt policy networks. We believe that this work is an important step forward towards creating generative dialog models that can not only generalize to large unlabelled datasets in complex domains but also be explainable to human users.

\section*{Acknowledgments}
This work was funded by NSF grant CNS- 1512973. The opinions expressed in this paper do not necessarily reflect those of NSF.

\bibliography{acl2018}
\bibliographystyle{acl_natbib}

\appendix
\section{Supplemental Material}
\subsection{Derivation}
\label{sec:anti-info-detail}
\begin{align}
    & \EX_x [\text{KL}(q(z|x) | p(z))] = \\ \nonumber
    & \EX_{q(z|x)p(x)} [\log(q(z|x)) - \log(p(z)))]\\ 
    & = - H(z|x) - \EX_{q(z)}[\log(p(z))] \\
    & = - H(z|x) + H(z) + \text{KL}(q(z)\|p(z)) \\
    & =  I(z, x) + \text{KL}(q(z)\| p(z))
\end{align}
where $q(z)=\EX_x[q(z|x)]$ and $I(z, x)=H(z)-H(z|x) $ is mutual information between $z$ and $x$ by definition.

\subsection{Interpolating Latent Space}
\label{sec:vae-interpolation}
Bowman et al.,~\shortcite{bowman2015generating} have shown that one can transform between two sentences by interpolating in the latent space of continuous VAEs. We found that our DI-VAE enjoys the same property. Specifically, two sentences $x_1$ and $x_2$ are sampled and their latent code are $q_{\recog}(z_1|x_1)$ and $q_{\recog}(z_2|x_2)$. We can then interpolate by flipping each latent code from $z^m_1$ to $z^m_2$, $m\in [1, M]$. For models with $M$ latent variables, one sentence can transform to another one in at most $M$ steps. Table~\ref{tbl:interpolating} shows an example. 
\begin{table}[ht]
\centering
\small
\begin{tabular}{l}
\textbf{So you can keep record of all the checks you write.}\\
So you can get all kinds of information and credit cards. \\
So you can keep track of all the credit cards. \\
So you kind of look at the credit union.\\
So you know of all the credit cards.\\
Yeah because you know of all the credit cards.\\
Right you know at least a lot of times.\\ 
\textbf{Right you know a lot of times.} \\
\end{tabular}
\label{tbl:interpolating}
\caption{Interpolating from the source sentence (top) to a target sentence (bottom) by sequentially setting the source latent code to the target code.}
\end{table}

\subsection{Data Details}
The details of the three dialog datasets are shown below.
\begin{table}[ht]
\centering
\small
\begin{tabular}{l|lll} \hline
                  & SMD      & DD                    & SW                    \\ \hline
Type              & Task     & Chat                  & Chat                  \\
Vocab Size        & 1,835    & 17,705                & 24,503  \\
\# of Dialogs     & 3,031    & 13,118                & 2,400                  \\
Avg Dialog Len    & 6.36     & 9.84                  & 59.2                  \\
Avg Utterance Len & 12.1     & 16.3                  & 22.1                  \\ \hline
\end{tabular}
\caption{Statistics of the dialog datasets. Vocabulary in DD and SW are capped to the most frequent 10K word types.}
\end{table}

\subsection{Training Details}
All RNNs use GRUs~\cite{chung2014empirical}. The GRU-RNNs for DI-VAE and DI-VST have hidden size 512. The utterance encoder in LAED has hidden size 256 for one direction and the context encoder and response decoder have hidden size 512. The word embedding is shared everywhere with embedding size 200. The temperature of Gumbel Softmax is set to 1. We train with Adams~\cite{kingma2014adam} with initial learning rate 0.001.
\end{document}